%% file: aaai2020.tex
\def\year{2020}\relax
\documentclass[letterpaper]{article} 
\usepackage{aaai20}  
\usepackage{times}  
\usepackage{helvet} 
\usepackage{courier}  
\usepackage[hyphens]{url}  
\usepackage{graphicx} 
\usepackage{graphicx}  
\usepackage{times}
\usepackage{latexsym}
\usepackage{amsmath}
\title{Simultaneously Linking Entities and Extracting Relations from \\ Biomedical Text Without Mention-level Supervision}

\author{Trapit Bansal$\dagger$ \and Pat Verga\thanks{work done while authors were at UMass Amherst}$\ddagger$ \and Neha Choudhary\footnotemark[1]$\dagger$ \and Andrew McCallum$\dagger$\\
$\dagger$University of Massachusetts, Amherst\\
$\dagger$\texttt{\{tbansal, nchoudhary, mccallum\}@cs.umass.edu}\\
$\ddagger$Google Research\\
$\ddagger$\texttt{patverga@google.com}
}

\newcommand{\citet}[1]{\citeauthor{#1} \shortcite{#1}} 
\newcommand{\citep}{\cite}

\begin{document}
\maketitle

\begin{abstract}
Understanding the meaning of text often involves reasoning about entities and their relationships. This requires identifying textual mentions of entities, linking them to a canonical concept, and discerning their relationships. These tasks are nearly always viewed as separate components within a pipeline, each requiring a distinct model and training data. While relation extraction can often be trained with readily available weak or distant supervision, entity linkers typically require expensive mention-level supervision -- which is not available in many domains. Instead, we propose a model which is trained to simultaneously produce entity linking and relation decisions while requiring no mention-level annotations. This approach avoids cascading errors that arise from pipelined methods and more accurately predicts entity relationships from text. We show that our model outperforms a state-of-the art entity linking and relation extraction pipeline on two biomedical datasets and can drastically improve the overall recall of the system. 

\end{abstract}

\section{Introduction \label{sec:intro}}

\begin{figure}[t]
\centering
\includegraphics[width=\linewidth]{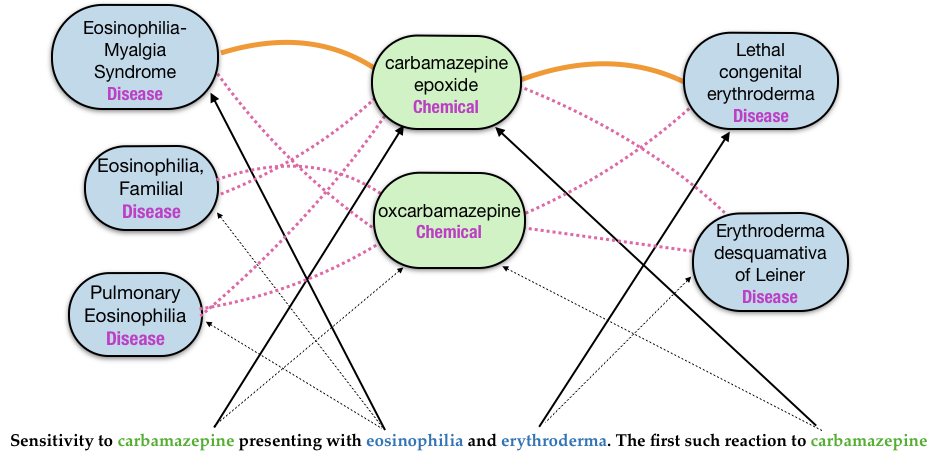}
\caption{Overview of the graph extraction task. Given a document represented as a title and abstract. Text mentions are denoted with color and each can link to one of several possible entities. The model considers the full set of entity linking and relation edges (all lines) and predicts the graph of true entities and relations (solid lines) represented in the text. Dashed lines show possible (incorrect) edges and solid lines show the true edges.}
\label{fig:task_overview}
\end{figure}

Making complex decisions in domains like biomedicine and clinical treatments requires access to information and facts in a form that can be easily viewed by experts and is computable by reasoning algorithms. The predominant paradigm for storing this type of data is in a knowledge graph. Much of these facts are populated from hand curation by human experts, inevitably leading to high levels of incompleteness \citep{bodenreider2004unified,bollacker2008freebase}. To address this, researchers have focused on automatically constructing knowledge bases by directly extracting information from text \citep{ji2010overview}. 

This procedure can be broken down into three major components; identifying mentions of entities in text \citep{ratinov2009design,lample2016neural,strubell2017fast}, linking mentions of the same entity together into a single canonical concept \citep{cucerzan2007large,gupta2017entity,raiman2018deeptype}, and identifying relationships occurring between those entities \citep{bunescu2007learning,wang2016relation,naacl18-verga}. 

These three stages are nearly always treated as separate serial components in an extraction pipeline and current state-of-the-art approaches train separate machine learning models for each component, each with their own distinct training data. More precisely, this data consists of mention-level supervision, that is individual instances of entities and relations which are identified and demarcated in text. This type of data can be prohibitively expensive to acquire, particularly in domains like biomedicine where expert knowledge is required to understand and annotate relevant information. 

In contrast, forms of distant supervision are readily available as database entries in existing knowledge bases. This type of information encodes global properties about entities and their relationships without identifying specific textual instances of those facts. This form of distant supervision has been successfully applied to relation extraction models \citep{mintz2009distant,surdeanu2012multi,riedel2013relation}. However, all of these methods still consume entity linking decisions as a preprocessing step, and unfortunately, accurate entity linkers and the mention-level supervision required to train them do not exist in many domains.

In this work, we instead develop a method to simultaneously link entities in the text and extract their relationships (see Fig.~\ref{fig:task_overview}). 
Our proposed method, called SNERL (\textbf{S}imultaneous \textbf{N}eural \textbf{E}ntity-\textbf{R}elation \textbf{L}inker), can be trained by leveraging readily available resources from existing knowledge bases and \textit{does not utilize any mention-level supervision}. In experiments performed on two different biomedical datasets, we show that our model is able to substantially outperform a state-of-the-art pipeline of entity linking and relation extraction by jointly training and testing the two tasks together.

\section{Methodology \label{sec:model}}

\begin{figure}[t]
\centering
\includegraphics[width=\linewidth]{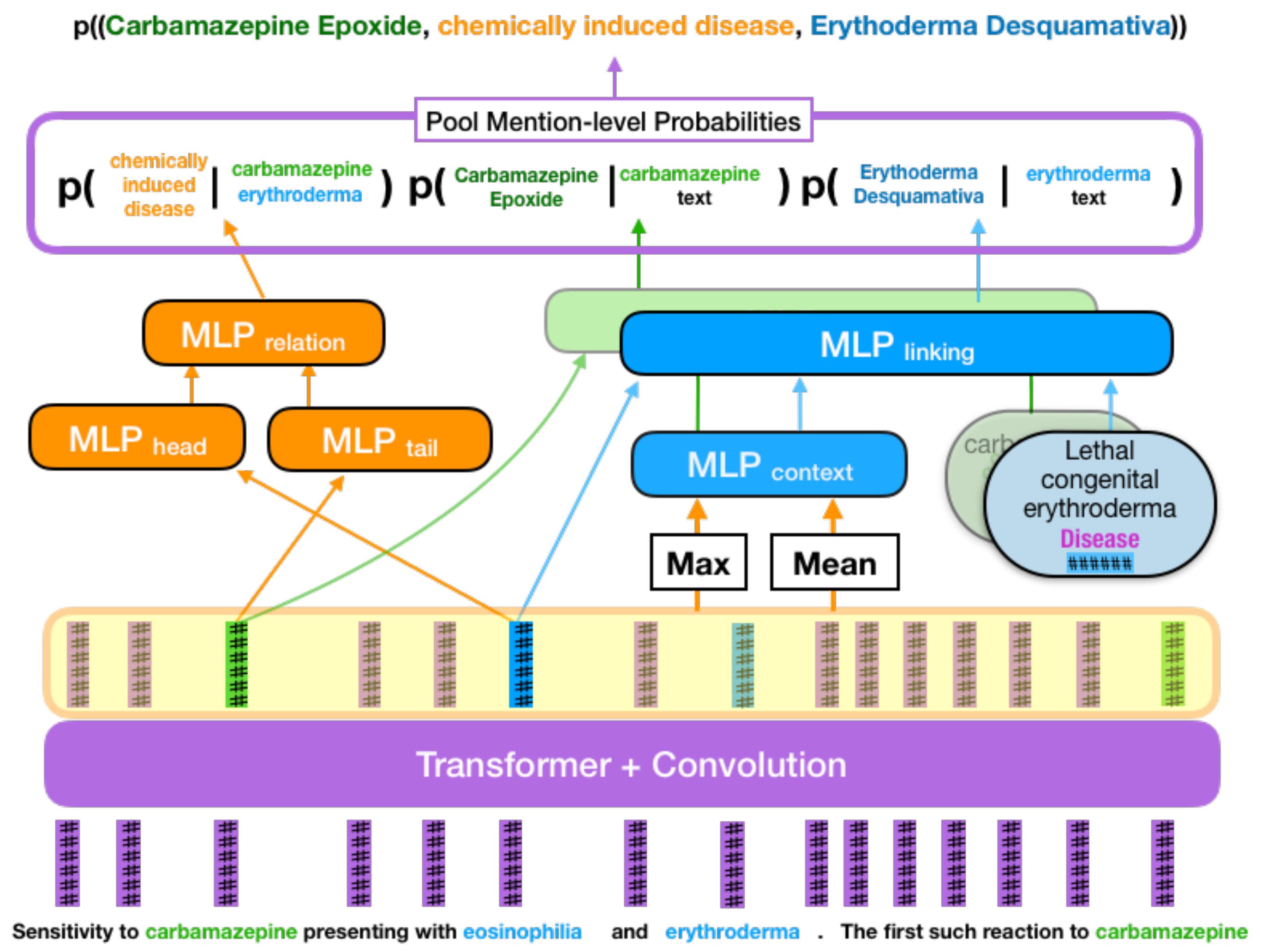}
\caption{Architecture of the SNERL model. The text of the title and abstract are mapped to word embeddings which is then contextually encoded using a transformer architecture. The left side of the figure shows the procedure for scoring an individual relation mention using a separate head and tail $\text{MLP}$ fed to a $\text{MLP}_{\text{relation}}$. The right side shows the entity linking component. The $\text{MLP}_{\text{linking}}$  model takes as input, an entity mention, a context representation derived from the mean and max over all contextualized token embeddings, and a candidate entity representation. These three probabilities (relation prediction and the two entity linking predictions) make up a single mention-level prediction. All mention-level predictions corresponding to the same entities are then pooled to make a final entity-level prediction.} \label{fig:model}
\end{figure}

In this section, we describe the proposed model, Simultaneous Neural Entity-Relation Linker (SNERL), and how it's trained. 
The input to the model is the full title and abstract of an article and the output is the predicted graph of entities and relations represented in the text (see Fig.~\ref{fig:task_overview}).
This is done by first encoding the text using self-attention \cite{vaswani2017attention} to obtain a contextualized representation of each entity mention in the input. These contextualized representations are then used to predict both the distribution over entities at the mention-level and the distribution over relations at the mention-pair-level. 
These predicted probabilities are then combined for each mention-pair and pooled at the document-level to get a final probability for predicting the tuple $(e_1, r, e_2)$ for the text (see Fig.~\ref{fig:model}).

\paragraph{Notations:} 
Let $[N]$ denote the set of natural numbers $\{1, \ldots, N\}$.
Each document consists of a set of words $\{x_i\}$ indexed by $i\in [V]$ where $V$ is the vocabulary size. 
Entity mentions in the document are found using a named entity recognition (NER) system \cite{wei2013pubtator}.
Let $\{m_j\}$ for $j \in [M]$ be the set of mention start indices for the document, where $M$ is the number of mentions in the document.
For each mention string $x_{m_i}$ we generate up to $\text{C}$ candidate entities (see Candidate Generation for details). 
Let $E$ be the set of all entities.
Each document is annotated with the graph of entities and relations, given as a set of tuples $G_d = \{(e_k, r, e_l)\}$, where $e_k, e_l \in E$ and $r \in [R]$. This is obtained from a knowledge base under the strong distant supervision assumption \cite{mintz2009distant} (see Experiments section for details).
Let $E_d \subset E$ be the set of entities in the annotations for the document $d$. 
$[a;b]$ denotes concatenation of vectors $a$ and $b$.

\subsection{Text Encoder}
The initial input to our model is the full title and abstract of a biomedical article from PubMed. 
\footnote{\url{https://www.ncbi.nlm.nih.gov/pubmed/}}
The sequence is tokenized and each token is mapped to a $n$-dimensional word embedding.
The sequence of word embeddings are the input to our text encoder.
The text encoder is based on the Transformer architecture of \citet{vaswani2017attention}. The transformer applies multiple blocks of multi-head self-attention followed by width 1 convolutions. We follow \citet{naacl18-verga} and add additional width 5 convolutions between blocks. The reader is referred to the Supplementary for the specific details. 
The text encoder, after multiple blocks of transformations, generates position and context informed hidden representations for each word in the document. 
The output of the text encoder is an $n$-dimensional contextualized embedding $h_i$ for each token $x_i$:
\begin{align*}
    h_1, \ldots, h_N &= \text{transformer}(x_1, \ldots, x_N)
\end{align*}
From an efficiency perspective, we only encode the document once and use the contextualized token representations to predict both the entities and the relations.

\subsection{Predicting entities \label{sec:model_linking}}
From the contextualized token representations $\{h_i\}$,
we first obtain a document representation by concatenating the mean-pooled and max-pooled token representations and projecting it through a multi-layer perceptron (MLP).
\begin{align*}
    \tilde{h} &= W^2_{\text{doc}}(\text{ReLU}(W^1_{\text{doc}} [\text{mean}(\{h_i\}); \text{max}(\{h_i\})] ) )
\end{align*}
where $\text{mean}(\cdot)$ denotes an element-wise mean of a set of vectors and $\text{max}(\cdot)$ denotes an element-wise max of a set of vectors.
Now, for each mention, we generate candidates entities for the mention. Such a candidate generation step is often used in entity-linking models \cite{shen2015entity} and in many domains, such as for Wikipedia entities, high quality candidates can be generated by using prior linking counts of mention surface forms to entities obtained from Wikipedia anchor texts \cite{ganea2017deep,raiman2018deeptype}. However, such high quality candidate generation is not available in the biomedical domain and so we resort to an approximate string matching approach for generating candidate entities.

\textbf{Candidate Generation}:
We followed procedures from previous work \citep{leaman2016taggerone,murty2018hierarchical}. Each mention was first normalized by removing all punctuation, lower-casing, and then stemming. Next, these strings were converted to tfidf vectors consisting of both word and character ngrams. We considered character ngrams of lengths two to five, and for words we considered unigrams and bigrams. The same procedure was also applied to convert all canonical string names and synonyms for entities in our knowledge base. Finally, candidates for each mention were generated according to their cosine similarity amongst all entities in the knowledge base.

For each candidate entity $e_i$ with type $t_i$, we generate a $n$-dimensional entity embedding as $\tilde{e}_i = \hat{e}_i + t_{i}$, by adding an entity-specific embedding $\hat{e}_i$ and a $n$-dimensional entity type embedding $t_{i}$. 
The entity-specific embedding can be learned or it can be a pre-trained embedding obtained from another source such as entity descriptions \cite{ganea2017deep,xie2016representation} or by a graph embedding method \cite{yang2014embedding}.
Now, for the $i$-th mention in the document, with starting index $m_i$, we consider $h_{m_i}$ as a contextualized mention representation and
define a score for predicting the candidate entity $e$ for this mention using the candidate representation $\tilde{e}$, document representation $\tilde{h}$, and mention representation $h_{m_i}$.
This is passed through a softmax function, normalizing over the set of candidates $C_{m_i}$ for the mention to get a probability $p(e | m_i, \text{text})$ for linking the mention $m_i$ to entity $e$.
\begin{align}
    l(e, m_i, \text{text}) &= W_l^2(\text{ReLU}(W_l^1[\tilde{e}; \tilde{h}; h_{m_i}]) \nonumber \\
    p(e | m_i, \text{text}) &= \underset{e \in C_{m_i}}{\text{softmax}} \left( l(e, m_i, \text{text}) \right) \label{eq:p_em}
\end{align}
We thus obtain a $(\text{M} \times \text{C})$ matrix of linking probabilities for the document, where $\text{M}$ is the maximum number of entity mentions in the document and $\text{C}$ is the maximum number of candidates per mention. 
\textit{Note that there is no direct mention-level supervision available to train these probabilities}.

\subsection{Predicting relations \label{sec:model_relations}}
Given the contextualized mention representation, we obtain a head and tail representation for each mention to serve as the head or tail entity of a relation tuple $(e_i, r, e_j)$. This is done by using two MLP to project each mention representation.
\begin{align*}
    e^{\text{head}}_{m_i} &= W_{\text{head}}^2(\text{ReLU}(W_{\text{head}}^1 h_{m_i})) \\
    e^{\text{tail}}_{m_j} &= W_{\text{tail}}^2(\text{ReLU}(W_{\text{tail}}^1 h_{m_j}))
\end{align*}
The head and tail representations are then passed through an MLP to predict a score for every relation $r$ for a pair of mentions $m_i$ and $m_j$. We pass this score vector through a sigmoid function to get a probability of predicting the relation from the mention-pair.
\begin{align}
    s(r, m_i, m_j) &= W_{r}^2(\text{ReLU}(W_{r}^1 [e^{\text{head}}_{m_i}; e^{\text{tail}}_{m_i}])) \nonumber \\
    p(r | m_i, m_j) &= \sigma (s(r, m_i, m_j)) \label{eq:p_r}
\end{align}
We thus obtain a $(\text{M} \times \text{M} \times \text{R})$ matrix of probabilities for predicting all relations, where $\text{R}$ is the maximum number of relations, from all pairs of entity mentions.

\subsection{Combining entity and relation predictions}
To predict the graph of entities and relations from the document, we need to assign a probability to every possible relation tuple $(e_k, r, e_l)$.
We first obtain the probability of predicting a tuple $(e_k, r, e_l)$ from a mention-pair $(m_i, m_j)$ by combining the probability for predicting the candidates for each of the mentions \eqref{eq:p_em} and the relation prediction probability \eqref{eq:p_r}. If an entity is not a candidate for a mention then it's entity prediction probability is zero for that mention.
\begin{align}
    &p\left((e_k, r, e_l) | m_i, m_j, \text{text} \right) = \nonumber \\
    &p(e_k | m_i, \text{text}) p(r | m_i, m_j) p(e_l | m_j, \text{text}) \label{eq:pt_m}
\end{align}
Then, the probability of extracting the tuple $(e_k, r, e_l)$ from the entire document can be obtained by pooling over all mention pairs $(m_i, m_j)$. For example, we can use max-pooling, which corresponds to the inductive bias that in order to extract a tuple we must find at least one mention pair for the corresponding entities in the document that is evidence for the tuple.
\begin{align}
    p\left((e_k, r, e_l) | \text{text} \right) &= \max_{i, j} p\left((e_k, r, e_l) | m_i, m_j, \text{text} \right) \label{eq:pt}
\end{align}

\paragraph{Soft maximum pooling:}
It has been observed previously \cite{verga2017generalizing,das2017chains} that the hard max operation is not ideal for pooling evidence as it leads to very sparse gradients. Recent methods, thus use the logsumexp function for pooling over \textit{logits}, which allows for more dense gradient updates.
However, we cannot use the logsumexp function in our case to pool over the probabilities \eqref{eq:pt_m} as the result of logsumexp over independent probabilities is not guaranteed to be a probability (in $[0, 1]$). Thus, we use a different operator that is considered a smooth relaxation of the maximum \cite{bansal2015content}. Given a set of elements $\{a_i\}$, the smooth-maximum (smax) with temperature $\tau$ is defined as:
\begin{align*}
    w_i &= \underset{i}{\text{softmax}} \left(\frac{a_i}{\tau}\right); \quad
    \textit{smax}(\{a_i\}) = \sum_i w_i a_i
\end{align*}
Note that for $\tau \rightarrow 0$ the result of \textit{smax} tends to the maximum of the set and for $\tau \rightarrow \infty$ the result is the average of the set. Thus, \textit{smax} can smoothly interpolate between these extremes. We use this \textit{smax} pooling over probabilities in \eqref{eq:pt} with a learned temperature $\tau$.

\subsection{Training}
We are given ground-truth annotation for the set of tuples in the document, $G_d = \{(e_k, r, e_l)\}$. 
We train based on the cross-entropy loss from predicted tuple probabilities \eqref{eq:pt}. 
Since we only have a subset of positive annotations, there is uncertainty in the set of negatives, and we deal with this by weighting the positive annotations by a weight $w_t$ in the cross-entropy loss. Let $y_{krl} = 1$ if document is annotated with the relation tuple $(e_k, r, e_l)$ and 0 otherwise, and $p_{krl}$ be its predicted probability in \eqref{eq:pt}, then we maximize $\log p(G_d | text)$:
\begin{align*}
    \frac{1}{|G_d|} \sum_{k,r,l} 
    w_t y_{krl} \log p_{krl} + (1-y_{krl}) \log (1 - p_{krl})
\end{align*}
In addition, since we can obtain \textit{document-level} entity annotations from the set of annotated relation tuples, we can provide an additional document-level entity supervision to better train our entity linking probabilities. To do this, we perform max-pooling over all mentions for each candidate entity for the document in \eqref{eq:p_em}, to obtain a document-level entity prediction score $p(e | \text{text}) = \max_m p(e | m, \text{text})$. We compute a weighted cross-entropy for these document-level predictions, again up-weighting the positive entities with a weight $w_e$. 
In summary, we combine graph prediction and document-level entity prediction objectives similar to multi-task learning \citep{caruana1993multitask}, so if $E_d$ is the set of entities in annotation, we maximize:
\begin{align}
    \log p(G_d | text) + \alpha \log p(E_d | text) \label{eq:loss}
\end{align}

Note that since we only have some positive annotations, there could be many mentions in the document for which the correct entity is not annotated. Thus, we down-weight the document-entity prediction term by $\alpha$ in the objective.  

\textbf{Technical Details:}
Since the size of $G_d$ can be very large, in order to improve training efficiency we subsample the set of unannotated entities as the negative entities to a maximum of $n^-$ per document.
Pooling over the joint mention-level probability \eqref{eq:pt} requires an intermediate $(\text{L} \times \text{L} \times \text{M} \times \text{M} \times \text{R})$ tensor, where $\text{L}$ is the total number of \textit{candidate} entities for the document. Since this can be computationally prohibitive, we compute the top-$k$ mentions per candidate entity based on the predicted probabilities \eqref{eq:p_em} and only backpropagate the gradients through the top-$k$. We consider $k$ as a hyperparameter and tune it on the validation set.

\section{Experiments \label{sec:ctd_results}}
Our experimental setting is that, for each test document (title and abstract), the model should predict the full graph of entity-relationships expressed in that document (a single example is depicted in Fig.~\ref{fig:task_overview}).
Thus, we evaluate on micro-averaged precision, recall and F1 for predicting \textit{the entire set of annotated relation tuples} across documents.
Our results show significant improvement in F1 over a state-of-the-art pipelined approach \cite{naacl18-verga}. Hyperparameters are in the Supplementary.

\subsection{Baselines}
All of our models use the same neural architecture described earlier, consume the same predicted entity mentions from an external NER model \citep{wei2013pubtator}, and \textit{differ} in how they produce entity linking decisions. The first two baselines take hard entity linking decisions as inputs and do not do any internal entity linking inference. Both these baselines are equivalent to the BRAN model from \citet{naacl18-verga} with two different ways of obtaining entity links for that model. This is a state-of-the-art \textit{pipelined approach} to entity-relation extraction.
We used an MLP as the relation scoring function for BRAN (similar to the SNERL model) as it performed better in experiments compared to the biaffine function used in the original paper. \\
\textbf{BRAN (Top Candidate)} 
produces entity linking decisions based on the highest scoring candidate entity (as described in `Candidate Generation' section).\\
\textbf{BRAN (Linker)} 
produces entity linking decisions from a trained state-of-the-art entity linker. We followed BRAN and obtained entity links from \citet{wei2013pubtator}.  \\
\textbf{SNERL} 
is our proposed model that does not take in any hard entity linking decisions as input and instead jointly predicts the full set of entities and relations within the text. For this model we considered 25 candidates per mention.

\subsection{CTD Dataset \label{sec:ctd_data}}
Our first set of experiments are on the CTD dataset first introduced in \citet{naacl18-verga}.
The data is derived from annotations in the Chemical Toxicology Database \citep{davis2018comparative}, a curated knowledge base containing relationships between chemicals, diseases, and genes. Each fact additionally contains a reference to the document (a scientific publication) where the annotator identified the relationship. Thus, these are used to obtain annotator identified entity-relationships in a given scientific publication. This type of document annotation is fairly common in biomedical knowledge bases, further motivating this work. This allows us to treat these annotations as a form of strong distant supervision \citep{mintz2009distant}. Here annotations are at the document-level rather than the mention-level (as in typical supervised learning) or corpus-level (as in standard distant supervision).


\begin{table}[t]
\centering
\begin{tabular}{lc}
\cline{1-2}
  & Tuple Recall    \\ 
\hline \hline
Top 1 Candidates & 67.0\% \\
Top 25 Candidates & \textbf{80.0\% } \\
Entity Linker & 60.4\% \\
\end{tabular}
\caption{Oracle recall for predicting entity-relation tuples under various models for selecting entity prediction, on the development set of CTD dataset. The oracle assumes perfect relation extraction recall. Note that to correctly extract a given entity-relation tuple, both head and tail entities in that relation need to predicted correctly. Since SNERL does not take entity links as input and has access to the top 25 candidates to make it's entity prediction, it can provide significantly higher recall.}
\label{tab:ctd_max recall}
\end{table}

An aspect of the document-level supervision is that the original facts were annotated over complete documents. However, due to paywalls we often only have access to titles and abstracts of papers. Therefore, there is no guarantee that the relationship is actually expressed in the title or abstract.
\citet{naacl18-verga}, thus, filtered the CTD dataset to only consider those entity-relation tuples where both entities are found in the text, for some mention, by the external entity linker. This ensures that all filtered tuples can be predicted by the model. However, this removes many correct entity-relationships that were indeed present but were filtered because those entities cannot be predicted by the entity linking model. 
We remedy this and create a more challenging train/development/test split from the entire CTD annotations, where we keep all entity-relationships in which the participating entities are a \textit{candidate} for some mention in the document. That is, for each annotated tuple $(e_1, r, e_2)$ between entities $e_1$ and $e_2$ in document $D$, we consider that tuple if both $e_1$ and $e_2$ are candidates for some mention in $D$.
Note that we used the candidate-generation approach described previously for generating the candidates for the mentions\footnote{we consider up to 250 candidates entities per mention for the data filtering}. Dataset statistics are in the Supplementary. We consider this as the Full CTD Dataset as it does not give an advantage to any particular entity linking model, but for completeness also evaluate on the subset filtered according to the original paper (we refer to this as BRAN-filtered).

To illustrate how the cascading errors of a pipelined approach of first predicting entity links and then predicting relations can degrade performance, we computed an oracle recall for the tuple prediction task on the development set of CTD. 
For this, we consider perfect accuracy on relation prediction, so the recall on tuple extraction is limited only by the entity linking accuracy. We consider three methods for entity linking: predicting the top candidate (based on the string similarity score from candidate generation), an oracle which can select the correct entity (if present) from the top 25 candidates and the trained entity liking model used by BRAN. Table \ref{tab:ctd_max recall} shows the results. We can see that errors from the entity linking step significantly restrict the models performance in pipelined approaches. On the other hand, if the model can infer the entity links (from top 25 candidates) jointly with the relations, it ameliorates this problem of cascading errors, potentilally leading to much higher recall. 


\begin{table}[t]
\centering
\begin{tabular}{llll}
\cline{1-4}
Model       & Precision & Recall & F1    \\ 
\hline \hline
BRAN (Top Candidate)   & 30.5 & 29.5  & 30.0 \\
BRAN (Linker)          & 33.2 & 28.1  & 30.5 \\
\textbf{SNERL}      & \textbf{41.1} & \textbf{43.4}  & \textbf{42.2}
\end{tabular}
\caption{Precision, Recall, and F1 for the full CTD test set. Bold values are statistically significant ($p$-value $<0.05$ using Wilcoxon signed-rank test) over the non-bold values in the same column.}
\label{tab:ctd_candidate_filtered}
\end{table}

\paragraph{Results on Full CTD data:} In table \ref{tab:ctd_candidate_filtered}, we can see that the SNERL model that jointly considers both entity and relations together drastically outperforms the models that take hard linking decisions from an external model. This is primarily due to huge drop in recall caused by cascading errors.

\paragraph{Results on BRAN Filtered CTD data:}
We also report results using the original filtering approach of \citet{naacl18-verga}.
Importantly, this approach gives a substantial advantage to the BRAN (Linker) baseline as the data is filtered to only consider the relationships for which it could potentially make a prediction. 
In table \ref{tab:ctd_link_filtered}, we can see that in spite of this disadvantage, the SNERL model is able to perform comparably to the BRAN (Linker) baseline.

\begin{table}[t]
\centering
\begin{tabular}{llll}
\cline{1-4}
Model       & Precision & Recall & F1    \\ 
\hline \hline
BRAN (Top Candidate) & 43.0    & 49.0  & 45.8 \\
BRAN (Linker)        & \textbf{45.7}    & 53.8  & \textbf{49.4} \\
\textbf{SNERL}    & \textbf{45.2}    & \textbf{55.2}  & \textbf{49.7}
\end{tabular}
\caption{Precision, Recall, and F1 for the BRAN-filtered CTD test data (i.e. filtered to tuples where BRAN can make a prediction). Bold values are statistically significant ($p$-value $<0.05$ using Wilcoxon signed-rank test) over the non-bold values in the same column, and the difference between multiple bold values in the same column is not statistically significant.  \label{tab:ctd_link_filtered}}
\end{table}

\subsection{CDR Entity Linking Performance \label{sec:cdr_linking_results}}
In order to evaluate how much of the success of the SNERL model can be attributed to the entity linking component \eqref{eq:p_em}, we evaluated its performance on the BioCreative V Chemical Disease Relation dataset (CDR) introduced in \citet{wei2015overview}. Similar to the CTD dataset, CDR was also originally derived from the Chemical Toxicology Database. Expert annotators chose 1,500 of those documents and exhaustively annotated all \textit{mentions} of chemicals and diseases in the text. Additionally, each mention was assigned its appropriate entity linking decision. We use this dataset as a gold standard to \textit{validate} our entity linking models. \textit{Note that we do not use this data for training, but only for evaluation}. 

We use the model that was trained on the CTD data and make it predict entities for every mention on the test set of CDR. We follow the standard practice of using the gold mention boundaries for evaluation only, to not confound the entity linking performance with mention-detection performance. 
In Table \ref{tab:cdr_linking}, we see that our SNERL does learn to link entities better than the top candidate. As is common when evaluating on this data, we consider document-level rather than mention-level entity linking evaluation \citep{leaman2016taggerone}, that is, how does the set of predicted entities compare to the gold set annotated in the document.
Note that the SNERL model additionally benefits from jointly predict entities and relations.
Breakdown of the results into Chemical and Disease prediction performance can be found in Supplementary.

\begin{table}[h]
\centering
\begin{tabular}{llll}
\cline{1-4}
Model       & Precision & Recall & F1    \\ 
\hline \hline
Top Candidate   & 79.0 & 86.8 & 82.7 \\ 
\textbf{SNERL} & \textbf{83.3} & \textbf{90.2} & \textbf{86.6}
\end{tabular}
\caption{Results for entity linking on the CDR dataset. Bold values are statistically significant ($p$-value $<0.05$ using Wilcoxon signed-rank test).} \label{tab:cdr_linking}
\end{table}

\subsection{Disease-Phenotype Relations \label{sec:phenotype_results}}

To further probe the performance of our model we created a dataset of disease / phenotype (aka symptom) relations. The goal here is to identify specific symptoms caused by a disease. This type of information is particularly important in clinical treatments as it can lead to earlier diagnosis of rare diseases, faster application of appropriate interventions, and better overall outcomes for patients. This task also serves to further motivate our methods as accurate entity linking models for phenotypes are not readily available, nor is sufficient mention-level training data to build a supervised classifier.

\textbf{Relation Annotations:}
We created this dataset with a similar technique to the construction of the CTD dataset. We started from the relations in the Human Phenotype Ontology \citep{kohler2018expansion} that were annotated with a document containing that relationship. 

\textbf{Mention Detection:}
For disease mention detection we followed the same procedure as CTD dataset and used the annotated mentions from \citet{wei2013pubtator}. Because there is not a readily available phenotype tagger, we trained our own model to identify mentions of phenotypes in text. We trained an iterated dilated convolution model \citet{strubell2017fast}.\footnote{\url{https://github.com/iesl/dilated-cnn-ner}} Our training data came from \citet{groza2015automatic}, which we split into train, dev, and test sets (see Supplementary). Our NER model achieved a micro F1 score of
72.57. 

We observed that disease and phenotype entity spans are often overlapping and nested. 
We thus over-generate the set of mentions by taking the predictions from both the taggers and adding them to the set of all mentions for the document, since our model is able to pool over all theses mentions even if they overlap.

\textbf{Entity Linking:}
We followed a similar procedure as described in section \ref{sec:model_linking} to generate phenotype entity linking candidates. Using the small set of gold entity linked text mentions from \citet{groza2015automatic} we were able to estimate our candidate's entity linking accuracy. 
Our top candidate achieved an accuracy of 46.8\% while the recall for 100 candidates was 76.5\%. This demonstrates the additional difficulty of the disease-phenotype dataset as these candidate accuracies are much lower than the results for CTD data.
See Supplementary Figure 1 for recall of the candidate set at different values of K. 

\citet{kohler2018expansion} annotations make use of several disease vocabularies from OMIM \citep{hamosh2005online}, ORHPANET \citep{pavan2017clinical} and DECIPHER \citep{bragin2013decipher} databases. For generating disease candidates, we use disease name strings from all of these. 
The external entity linker that we used from \citet{wei2013pubtator} links diseases to the MeSH disease vocabulary. To align these with our disease-phenotype relation annotations, we use the MEDIC database \citep{davis2012medic} for mapping OMIM disease terms into the MeSH vocabulary.

The final dataset annotations were selected by filtering based on entities that can be found in document when considering up to 250 candidates per mention. See Supplementary for dataset statistics.




\subsubsection{Pre-training Entity Embedding \label{sec:phenotype_details}}
Since the dataset has many unseen entities at test time, we need a method to address these unseen entities as generating the linking probabilities in \eqref{eq:p_em} requires an entity embedding.
For this, we obtained entity descriptions for the phenotypes and encoded them using pre-trained sentence embedding from BioSentVec \cite{chen2018biosentvec}.
However, not all test entities have descriptions. So, in addition to the descriptions we trained a graph embedding model, DistMult \cite{yang2014embedding}, on the graph obtained from the set of all annotations in Human Phenotype Ontology excluding the dev/test annotations.
We project both these pre-trained embeddings using a learned linear transformation and sum the description and graph embedding to obtain the entity-specific embedding $\hat{e}$.

\subsubsection{Baselines}
We use the same baselines as before. For BRAN (Linker), the disease entity links come \citet{wei2013pubtator} and since we don't have access to an accurate pretrained phenotype entity linking model, this model also uses the top phenotype candidate as a hard phenotype entity linking decision.

\subsubsection{Results}

Our disease-phenotype results show a similar trend to those from the CTD experiments. Overall, the BRAN (Top Candidate) model performs the worst and the SNERL model outperforms both models that use hard entity linking decisions. 

Overall, our results indicate that this particular task is extremely challenging. This is likely the combination of several difficulties. The first is that the candidate set itself is not as accurate as the ones from the CTD experiment which we can see from comparing the top candidate accuracy of 46.8\% with the Top Candidate results in Table \ref{tab:cdr_linking}. Since we rely on the candidate set to filter the annotations for the documents, we might end up with significant annotations that are not present in the title and abstract.  Secondly, the amount of training data is significantly less (see Supplementary) than in the CTD experiments, requiring research into unsupervised approaches \cite{devlin2018bert} for this data. Lastly, dealing with out-of-vocabulary entities at test time required additional pre-training, and our analysis indicated that these are not highly predictive for mention-level disambiguation due to the sparsity of the graph training data. Looking into more sophisticated embedding methods \cite{xie2016representation,gupta2017entity,bansal2019a2n} and methods for dealing with unseen entities would be an important problem for future work.

\begin{table}[t]
\centering
\begin{tabular}{llll}
\cline{1-4}
Model       & Precision & Recall & F1    \\ 
\hline \hline
BRAN (Top Candidate) & 8.9 & 5.3 & 6.6 \\
BRAN (Linker) & 11.3 & 6.6  & 8.3 \\
\textbf{SNERL}  & \textbf{12.8} & \textbf{10.9}  & \textbf{11.8}
\end{tabular}
\caption{Results on the disease phenotype dataset \label{tab:disease_phenotype}. Bold values are statistically significant ($p$-value $<0.05$ using Wilcoxon signed-rank test) over non-bold values.}
\end{table}

\section{Related Work \label{sec:related_Work}}
Extracting entities and relations from text has been widely studied over the past few decades. In the biomedical domain specifically, there has been substantial progress on entity mention detection \cite{greenberg2018marginal,wei2015gnormplus} and entity linking (often referred to as normalization in the bio NLP community) \citep{leaman2008banner,leaman2013dnorm,leaman2015tmchem,leaman2016taggerone}, and relation extraction \citep{wei2016assessing,krallinger2017overview}. 
There have also been numerous works that have identified both entity mentions and relationships from text in both the general domain \citep{miwa2016end} and in the biomedical domain \citep{li2017neural,ammar2017ai2,naacl18-verga}. \citet{leaman2016taggerone} showed that jointly considering named entity recognition (NER) and linking led to improved performance. 

A few works have shown that jointly modeling relations and entity linking can improve performance. \citet{le2018improving} improved entity linking performance by modeling latent relations between entities. This is similar to coherence models \cite{ganea2017deep} in entity linking which consider the joint assignment of all linking decisions, but is more tractable as it focuses on only pairs of entities in a short context.
\citet{luan2018multi} created a multi-task learning model for predicting entities, relations, and coreference in scientific documents. This model required supervision for all three tasks and predictions amongst the different tasks were made independently rather than jointly.
To the best of our knowledge, SNERL is the first model that simultaneously links entities and predicts relations without requiring expensive mention-level annotation. 

\section{Conclusion}
In this paper, we presented a novel method, SNERL, to simultaneously predict entity linking and entity relation decisions. SNERL can be trained without any mention-level supervision for entities or relations, and instead relies solely on weak and distant supervision at the document-level, readily available in many biomedical knowledge bases. The proposed model performs favorably as compared to a state-of-the-art pipeline approach to relation extraction by avoiding cascading errors, while requiring less expensive annotation, opening possibilities for knowledge extraction in low-resource and expensive to annotate domains. 


\section*{Acknowledgments}
We thank Andrew Su and Haw-Shiuan Chang for early discussions on the disease-phenotype task.
This work was supported in part by the UMass Amherst Center for Data Science and the Center for Intelligent Information Retrieval, in part by the Chan Zuckerberg Initiative under the project Scientific Knowledge Base Construction, and in part by the National Science Foundation under Grant No. IIS-1514053. Any opinions, findings and conclusions or recommendations expressed in this material are those of the authors and do not necessarily reflect those of the sponsor.


{\fontsize{9.1pt}{10.1pt} \selectfont
\bibliography{references}}
\bibliographystyle{aaai}

\appendix
\section{\LARGE {\sc Supplementary Material}}
\input{Supplementary/appendix.tex}

\end{document}

%% file: Supplementary/appendix.tex
\section{Transformer Text Encoder}
The model takes a sequence of N word embeddings as input, $\{x_1, \ldots, x_N\}$.
Since the Transformer has no innate notion of position, the model relies on positional embeddings which are added to the
input token embeddings. The positional embeddings are also learned parameters of the model. Thus, we get the token representations:
\begin{equation*}
    s_i = x_i + p_i
\end{equation*}
Transformer \citep{vaswani2017attention} is made up of B blocks. Each Transformer block, denoted transformer$_k$, has its own set of parameters and consists of two components: multi-head attention followed by a series of convolutions. The output for token $i$ of block $k$, $b_i^k$, is connected to its input $b_i^{k-1}$ with a residual connection:
\begin{equation}
    b_i^k = b_i^{k-1} + \mbox{transformer}_k(b_i^{k-1}) \label{eq:block}
\end{equation}
In each block, multi-head attention \citep{vaswani2017attention} applies self-attention multiple times over the same inputs using separately normalized parameters (attention heads) and combines the results. Each head in multi-head self-attention updates its input $b_i^{k-1}$ by performing a weighted sum over all tokens in the sequence, weighted by their importance for modeling token $i$. Refer to \citet{vaswani2017attention} for details of multi-head attention.
The outputs of the individual attention heads are concatenated, to give the output of multi-head attention at the $i$-th token, $o_i = [o_{i1}\; \ldots; o_{in_h}]$. This is followed by layer normalization \citep{ba2016layer}, and two width-1 convolutions. Following \citep{naacl18-verga}, we add a third layer with kernel width 5 convolutions, which allows explicit n-gram modeling useful for relation extraction. This gives the output at the $i$-th token for the $k$-th transformer block in \eqref{eq:block}.

The sequence of representations at each token obtained after $B$ blocks of processing, described above, is the final output of the transformer text encoder:
\begin{equation*}
   \mbox{transformer}(x_1, \ldots, x_N) = h_1, \ldots, h_N = b_1^B, \ldots, b_N^B
\end{equation*}

\section{CTD Dataset}
The number of documents in the splits of CTD dataset are given in Table \ref{tab:ctd}.
There are 19933 entities and 14 relation types in this data. 
The number of entity-relationship tuples in train/dev/test are given in Table \ref{tab:ctd_tuples}.
\begin{table}[ht]
\centering
\begin{tabular}{llll}
\cline{1-4}
Data       & Train & Dev & Test    \\ 
\cline{1-4}
Full CTD   & 52,003 & 8,177 & 8,284 \\
BRAN-filtered CTD  & 45,586 & 5,857 & 5,804
\end{tabular}
\caption{Number of documents in CTD dataset\label{tab:ctd}}
\end{table}

\begin{table}[ht]
\centering
\begin{tabular}{llll}
\cline{1-4}
Data       & Train & Dev & Test    \\ 
\cline{1-4}
Full CTD   & 140,121 & 34,213 & 36,656 \\
BRAN-filtered CTD  & 115,319 & 14,141 & 14,372
\end{tabular}
\caption{Number of entity-relationship tuples in CTD dataset\label{tab:ctd_tuples}}
\end{table}

\section{Entity Linking on CDR}
Table \ref{tab:cdr_dis} shows the entity linking performance for diseases on the CDR dataset.
Table \ref{tab:cdr_chem} shows the entity linking performance for diseases on the CDR dataset.

\begin{table}[ht]
\centering
\begin{tabular}{llll}
\cline{1-4}
Model       & Precision & Recall & F1    \\ 
\cline{1-4}
Top Candidate   & 77.3 & 80.8 & 79.0 \\
SNERL      & \textbf{83.6} & \textbf{86.0} & \textbf{84.8}
\end{tabular}
\caption{Disease entity linking on the CDR dataset \label{tab:cdr_dis}}
\end{table}

\begin{table}[ht]
\centering
\begin{tabular}{llll}
\cline{1-4}
Model       & Precision & Recall & F1    \\ 
\cline{1-4}
Top Candidate   & 81.4 & 95.6 & 87.9 \\
SNERL      & \textbf{83.0} & \textbf{96.5} & \textbf{89.3}
\end{tabular}
\caption{Chemical entity linking on the CDR dataset \label{tab:cdr_chem}}
\end{table}

\section{Implementation Details \label{sec:implementation}}
All word embeddings are randomly initialized. Text is tokenized using the Genia tokenizer \citep{kulick2004integrated}.
We used dropout \cite{srivastava2014dropout} at the input word embeddings ($p_i$), on attention weights in the transformer \cite{vaswani2017attention} ($p_t$), after head and tail projection MLP ($p_s$), and after the first layers of the relation ($p_s$) and linking MLP ($p_s$). 
We also apply dropout to the input words replacing words with a special UNK token ($p_w$). Note that the values of dropout $p_*$ reported here are keep probabilities.
We used Adam \cite{kingma2014adam} for optimization with a learning rate of 0.001.
We tuned the dropout rates, weights for the cross-entropy term $w_t, w_e$, number of blocks $B$ of transformer, number of heads $n_h$, number of negative samples $n^-$, $k$ for the number of top mentions, and the weight $\alpha$ for the objective. 

On CTD, full dataset, the best hyperparameters were: $p_i = 0.25$, $p_t = 0.25$, $p_s = .15$, $p_w = 0.2$, $w_t = 5.0$, $w_e = 2.0$, $n^-=100$, $B=4$, $n_h=2$ $k=15$, $\alpha = 0.1$. We used embedding dimension $n=128$. 

On CTD, BRAN-filtered dataset, the best hyperparameters were: $p_i = 0.2$, $p_t = 0.$, $p_s = 0.$, $p_w = 0.3$, $w_t = 5.0$, $w_e = 0.$, $n^-=200$, $B=2$, $n_h=8$ $k=10$, $\alpha = 0.$. We used embedding dimension $n=128$.

On the Disease-Phenotype dataset, the best hyperparameters were: $p_i = 0.4$, $p_t = 0.4$, $p_s = 0.15$, $p_w = 0.15$, $w_t = 5.0$, $w_e = 0.$, $n^-=400$, $B=4$, $n_h=4$ $k=10$, $\alpha = 0.$. We used embedding dimension $n=64$.

\begin{figure}[ht]
\centering
\includegraphics[scale=.35]{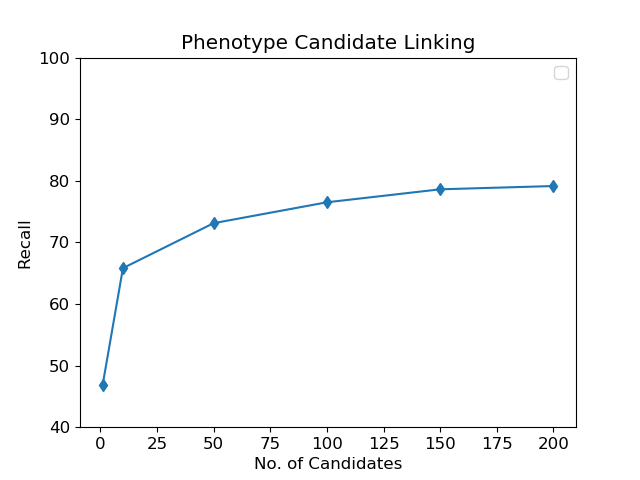}
\caption{Recall for different numbers of candidates for phenotype entity linking  \label{fig:phenotype_lnking}}
\end{figure}

\section{Disease-Phenotype Dataset}
Our final NER model achieved a micro F1 score of 75.00 on the development set and 72.57. 
The train/dev/test split consisted of 173/23/23 documents with 1294/118/160 metions respectively.

Table \ref{tab:phenotype_relex_stats} shows the train/dev/test splits for the disease-phenotype relation extraction dataset.

Figure \ref{fig:phenotype_lnking} shows the recall$@k$ for phenotype candidate set.


\begin{table}[t!]
\centering
\begin{tabular}{lll}
\cline{1-3}
Split       & Docs & Relations    \\ 
\cline{1-3}
Train & 401 & 1631 \\
Dev & 86 & 303 \\
Test  & 86 & 455 \\
\end{tabular}
\caption{Statistics for disease phenotype relation data \label{tab:phenotype_relex_stats}}
\end{table}

%% file: aaai2020.bbl
\begin{thebibliography}{}

\bibitem[\protect\citeauthoryear{Ammar \bgroup et al\mbox.\egroup
  }{2017}]{ammar2017ai2}
Ammar, W.; Peters, M.; Bhagavatula, C.; and Power, R.
\newblock 2017.
\newblock The ai2 system at semeval-2017 task 10 (scienceie): semi-supervised
  end-to-end entity and relation extraction.
\newblock In {\em Proceedings of the 11th International Workshop on Semantic
  Evaluation (SemEval-2017)},  592--596.

\bibitem[\protect\citeauthoryear{Ba, Kiros, and Hinton}{2016}]{ba2016layer}
Ba, J.~L.; Kiros, J.~R.; and Hinton, G.~E.
\newblock 2016.
\newblock Layer normalization.
\newblock {\em arXiv preprint arXiv:1607.06450}.

\bibitem[\protect\citeauthoryear{Bansal \bgroup et al\mbox.\egroup
  }{2019}]{bansal2019a2n}
Bansal, T.; Juan, D.-C.; Ravi, S.; and McCallum, A.
\newblock 2019.
\newblock A2n: Attending to neighbors for knowledge graph inference.
\newblock In {\em Proceedings of the 57th Annual Meeting of the Association for
  Computational Linguistics},  4387--4392.

\bibitem[\protect\citeauthoryear{Bansal, Das, and
  Bhattacharyya}{2015}]{bansal2015content}
Bansal, T.; Das, M.; and Bhattacharyya, C.
\newblock 2015.
\newblock Content driven user profiling for comment-worthy recommendations of
  news and blog articles.
\newblock In {\em Proceedings of the 9th ACM Conference on Recommender
  Systems},  195--202.
\newblock ACM.

\bibitem[\protect\citeauthoryear{Bodenreider}{2004}]{bodenreider2004unified}
Bodenreider, O.
\newblock 2004.
\newblock The unified medical language system (umls): integrating biomedical
  terminology.
\newblock {\em Nucleic acids research} 32(suppl\_1):D267--D270.

\bibitem[\protect\citeauthoryear{Bollacker \bgroup et al\mbox.\egroup
  }{2008}]{bollacker2008freebase}
Bollacker, K.; Evans, C.; Paritosh, P.; Sturge, T.; and Taylor, J.
\newblock 2008.
\newblock Freebase: a collaboratively created graph database for structuring
  human knowledge.
\newblock In {\em Proceedings of the 2008 ACM SIGMOD international conference
  on Management of data},  1247--1250.
\newblock AcM.

\bibitem[\protect\citeauthoryear{Bragin \bgroup et al\mbox.\egroup
  }{2013}]{bragin2013decipher}
Bragin, E.; Chatzimichali, E.~A.; Wright, C.~F.; Hurles, M.~E.; Firth, H.~V.;
  Bevan, A.~P.; and Swaminathan, G.~J.
\newblock 2013.
\newblock Decipher: database for the interpretation of phenotype-linked
  plausibly pathogenic sequence and copy-number variation.
\newblock {\em Nucleic acids research} 42(D1):D993--D1000.

\bibitem[\protect\citeauthoryear{Bunescu and
  Mooney}{2007}]{bunescu2007learning}
Bunescu, R., and Mooney, R.
\newblock 2007.
\newblock Learning to extract relations from the web using minimal supervision.
\newblock In {\em Proceedings of the 45th Annual Meeting of the Association of
  Computational Linguistics},  576--583.

\bibitem[\protect\citeauthoryear{Caruana}{1993}]{caruana1993multitask}
Caruana, R.
\newblock 1993.
\newblock Multitask learning: A knowledge-based source of inductive bias.
\newblock {\em Proceedings of the Tenth International Conference on
  International Conference on Machine Learning (ICML)}  41--48.

\bibitem[\protect\citeauthoryear{Chen, Peng, and Lu}{2018}]{chen2018biosentvec}
Chen, Q.; Peng, Y.; and Lu, Z.
\newblock 2018.
\newblock Biosentvec: creating sentence embeddings for biomedical texts.
\newblock {\em arXiv preprint arXiv:1810.09302}.

\bibitem[\protect\citeauthoryear{Cucerzan}{2007}]{cucerzan2007large}
Cucerzan, S.
\newblock 2007.
\newblock Large-scale named entity disambiguation based on wikipedia data.
\newblock In {\em Proceedings of the 2007 Joint Conference on Empirical Methods
  in Natural Language Processing and Computational Natural Language Learning
  (EMNLP-CoNLL)}.

\bibitem[\protect\citeauthoryear{Das \bgroup et al\mbox.\egroup
  }{2017}]{das2017chains}
Das, R.; Neelakantan, A.; Belanger, D.; and McCallum, A.
\newblock 2017.
\newblock Chains of reasoning over entities, relations, and text using
  recurrent neural networks.
\newblock In {\em Proceedings of the 15th Conference of the European Chapter of
  the Association for Computational Linguistics}, volume~1,  132--141.

\bibitem[\protect\citeauthoryear{Davis \bgroup et al\mbox.\egroup
  }{2012}]{davis2012medic}
Davis, A.~P.; Wiegers, T.~C.; Rosenstein, M.~C.; and Mattingly, C.~J.
\newblock 2012.
\newblock Medic: a practical disease vocabulary used at the comparative
  toxicogenomics database.
\newblock {\em Database} 2012.

\bibitem[\protect\citeauthoryear{Davis \bgroup et al\mbox.\egroup
  }{2018}]{davis2018comparative}
Davis, A.~P.; Grondin, C.~J.; Johnson, R.~J.; Sciaky, D.; McMorran, R.;
  Wiegers, J.; Wiegers, T.~C.; and Mattingly, C.~J.
\newblock 2018.
\newblock The comparative toxicogenomics database: update 2019.
\newblock {\em Nucleic acids research} 47(D1):D948--D954.

\bibitem[\protect\citeauthoryear{Devlin \bgroup et al\mbox.\egroup
  }{2018}]{devlin2018bert}
Devlin, J.; Chang, M.-W.; Lee, K.; and Toutanova, K.
\newblock 2018.
\newblock Bert: Pre-training of deep bidirectional transformers for language
  understanding.
\newblock {\em arXiv preprint arXiv:1810.04805}.

\bibitem[\protect\citeauthoryear{Ganea and Hofmann}{2017}]{ganea2017deep}
Ganea, O.-E., and Hofmann, T.
\newblock 2017.
\newblock Deep joint entity disambiguation with local neural attention.
\newblock {\em arXiv preprint arXiv:1704.04920}.

\bibitem[\protect\citeauthoryear{Greenberg \bgroup et al\mbox.\egroup
  }{2018}]{greenberg2018marginal}
Greenberg, N.; Bansal, T.; Verga, P.; and McCallum, A.
\newblock 2018.
\newblock Marginal likelihood training of bilstm-crf for biomedical named
  entity recognition from disjoint label sets.
\newblock In {\em Proceedings of the 2018 Conference on Empirical Methods in
  Natural Language Processing},  2824--2829.

\bibitem[\protect\citeauthoryear{Groza \bgroup et al\mbox.\egroup
  }{2015}]{groza2015automatic}
Groza, T.; K{\"o}hler, S.; Doelken, S.; Collier, N.; Oellrich, A.; Smedley, D.;
  Couto, F.~M.; Baynam, G.; Zankl, A.; and Robinson, P.~N.
\newblock 2015.
\newblock Automatic concept recognition using the human phenotype ontology
  reference and test suite corpora.
\newblock {\em Database} 2015.

\bibitem[\protect\citeauthoryear{Gupta, Singh, and
  Roth}{2017}]{gupta2017entity}
Gupta, N.; Singh, S.; and Roth, D.
\newblock 2017.
\newblock Entity linking via joint encoding of types, descriptions, and
  context.
\newblock In {\em Proceedings of the 2017 Conference on Empirical Methods in
  Natural Language Processing},  2681--2690.

\bibitem[\protect\citeauthoryear{Hamosh \bgroup et al\mbox.\egroup
  }{2005}]{hamosh2005online}
Hamosh, A.; Scott, A.~F.; Amberger, J.~S.; Bocchini, C.~A.; and McKusick, V.~A.
\newblock 2005.
\newblock Online mendelian inheritance in man (omim), a knowledgebase of human
  genes and genetic disorders.
\newblock {\em Nucleic acids research} 33(suppl\_1):D514--D517.

\bibitem[\protect\citeauthoryear{Ji \bgroup et al\mbox.\egroup
  }{2010}]{ji2010overview}
Ji, H.; Grishman, R.; Dang, H.~T.; Griffitt, K.; and Ellis, J.
\newblock 2010.
\newblock Overview of the tac 2010 knowledge base population track.
\newblock In {\em Third Text Analysis Conference (TAC 2010)}, volume~3,  3--3.

\bibitem[\protect\citeauthoryear{Kingma and Ba}{2014}]{kingma2014adam}
Kingma, D.~P., and Ba, J.
\newblock 2014.
\newblock Adam: A method for stochastic optimization.
\newblock {\em arXiv preprint arXiv:1412.6980}.

\bibitem[\protect\citeauthoryear{K{\"o}hler \bgroup et al\mbox.\egroup
  }{2018}]{kohler2018expansion}
K{\"o}hler, S.; Carmody, L.; Vasilevsky, N.; Jacobsen, J. O.~B.; Danis, D.;
  Gourdine, J.-P.; Gargano, M.; Harris, N.~L.; Matentzoglu, N.; McMurry, J.~A.;
  et~al.
\newblock 2018.
\newblock Expansion of the human phenotype ontology (hpo) knowledge base and
  resources.
\newblock {\em Nucleic acids research} 47(D1):D1018--D1027.

\bibitem[\protect\citeauthoryear{Krallinger \bgroup et al\mbox.\egroup
  }{2017}]{krallinger2017overview}
Krallinger, M.; Rabal, O.; Akhondi, S.~A.; et~al.
\newblock 2017.
\newblock Overview of the biocreative vi chemical-protein interaction track.
\newblock In {\em Proceedings of the sixth BioCreative challenge evaluation
  workshop}, volume~1,  141--146.

\bibitem[\protect\citeauthoryear{Kulick \bgroup et al\mbox.\egroup
  }{2004}]{kulick2004integrated}
Kulick, S.; Bies, A.; Liberman, M.; Mandel, M.; McDonald, R.; Palmer, M.;
  Schein, A.; Ungar, L.; Winters, S.; and White, P.
\newblock 2004.
\newblock Integrated annotation for biomedical information extraction.
\newblock In {\em HLT-NAACL 2004 Workshop: Linking Biological Literature,
  Ontologies and Databases}.

\bibitem[\protect\citeauthoryear{Lample \bgroup et al\mbox.\egroup
  }{2016}]{lample2016neural}
Lample, G.; Ballesteros, M.; Subramanian, S.; Kawakami, K.; and Dyer, C.
\newblock 2016.
\newblock Neural architectures for named entity recognition.
\newblock {\em arXiv preprint arXiv:1603.01360}.

\bibitem[\protect\citeauthoryear{Le and Titov}{2018}]{le2018improving}
Le, P., and Titov, I.
\newblock 2018.
\newblock Improving entity linking by modeling latent relations between
  mentions.
\newblock In {\em Proceedings of the 56th Annual Meeting of the Association for
  Computational Linguistics}, volume~1,  1595--1604.

\bibitem[\protect\citeauthoryear{Leaman and Gonzalez}{2008}]{leaman2008banner}
Leaman, R., and Gonzalez, G.
\newblock 2008.
\newblock Banner: an executable survey of advances in biomedical named entity
  recognition.
\newblock In {\em Biocomputing 2008}. World Scientific.
\newblock  652--663.

\bibitem[\protect\citeauthoryear{Leaman and Lu}{2016}]{leaman2016taggerone}
Leaman, R., and Lu, Z.
\newblock 2016.
\newblock Taggerone: joint named entity recognition and normalization with
  semi-markov models.
\newblock {\em Bioinformatics} 32(18):2839--2846.

\bibitem[\protect\citeauthoryear{Leaman, Islamaj~Do{\u{g}}an, and
  Lu}{2013}]{leaman2013dnorm}
Leaman, R.; Islamaj~Do{\u{g}}an, R.; and Lu, Z.
\newblock 2013.
\newblock Dnorm: disease name normalization with pairwise learning to rank.
\newblock {\em Bioinformatics} 29(22):2909--2917.

\bibitem[\protect\citeauthoryear{Leaman, Wei, and Lu}{2015}]{leaman2015tmchem}
Leaman, R.; Wei, C.-H.; and Lu, Z.
\newblock 2015.
\newblock tmchem: a high performance approach for chemical named entity
  recognition and normalization.
\newblock {\em Journal of cheminformatics} 7(1):S3.

\bibitem[\protect\citeauthoryear{Li \bgroup et al\mbox.\egroup
  }{2017}]{li2017neural}
Li, F.; Zhang, M.; Fu, G.; and Ji, D.
\newblock 2017.
\newblock A neural joint model for entity and relation extraction from
  biomedical text.
\newblock {\em BMC bioinformatics} 18(1):198.

\bibitem[\protect\citeauthoryear{Luan \bgroup et al\mbox.\egroup
  }{2018}]{luan2018multi}
Luan, Y.; He, L.; Ostendorf, M.; and Hajishirzi, H.
\newblock 2018.
\newblock Multi-task identification of entities, relations, and coreference for
  scientific knowledge graph construction.
\newblock In {\em Proceedings of the 2018 Conference on Empirical Methods in
  Natural Language Processing},  3219--3232.

\bibitem[\protect\citeauthoryear{Mintz \bgroup et al\mbox.\egroup
  }{2009}]{mintz2009distant}
Mintz, M.; Bills, S.; Snow, R.; and Jurafsky, D.
\newblock 2009.
\newblock Distant supervision for relation extraction without labeled data.
\newblock In {\em Proceedings of the Joint Conference of the 47th Annual
  Meeting of the ACL and the 4th International Joint Conference on Natural
  Language Processing of the AFNLP},  1003--1011.
\newblock Association for Computational Linguistics.

\bibitem[\protect\citeauthoryear{Miwa and Bansal}{2016}]{miwa2016end}
Miwa, M., and Bansal, M.
\newblock 2016.
\newblock End-to-end relation extraction using lstms on sequences and tree
  structures.
\newblock In {\em Proceedings of the 54th Annual Meeting of the Association for
  Computational Linguistics}, volume~1,  1105--1116.

\bibitem[\protect\citeauthoryear{Murty \bgroup et al\mbox.\egroup
  }{2018}]{murty2018hierarchical}
Murty, S.; Verga, P.; Vilnis, L.; Radovanovic, I.; and McCallum, A.
\newblock 2018.
\newblock Hierarchical losses and new resources for fine-grained entity typing
  and linking.
\newblock In {\em Proceedings of the 56th Annual Meeting of the Association for
  Computational Linguistics}, volume~1,  97--109.

\bibitem[\protect\citeauthoryear{Pavan \bgroup et al\mbox.\egroup
  }{2017}]{pavan2017clinical}
Pavan, S.; Rommel, K.; Marquina, M. E.~M.; H{\"o}hn, S.; Lanneau, V.; and Rath,
  A.
\newblock 2017.
\newblock Clinical practice guidelines for rare diseases: the orphanet
  database.
\newblock {\em PloS one} 12(1):e0170365.

\bibitem[\protect\citeauthoryear{Raiman and Raiman}{2018}]{raiman2018deeptype}
Raiman, J.~R., and Raiman, O.~M.
\newblock 2018.
\newblock Deeptype: multilingual entity linking by neural type system
  evolution.
\newblock In {\em Thirty-Second AAAI Conference on Artificial Intelligence}.

\bibitem[\protect\citeauthoryear{Ratinov and Roth}{2009}]{ratinov2009design}
Ratinov, L., and Roth, D.
\newblock 2009.
\newblock Design challenges and misconceptions in named entity recognition.
\newblock In {\em Proceedings of the thirteenth conference on computational
  natural language learning},  147--155.
\newblock Association for Computational Linguistics.

\bibitem[\protect\citeauthoryear{Riedel \bgroup et al\mbox.\egroup
  }{2013}]{riedel2013relation}
Riedel, S.; Yao, L.; McCallum, A.; and Marlin, B.~M.
\newblock 2013.
\newblock Relation extraction with matrix factorization and universal schemas.
\newblock In {\em Proceedings of the 2013 Conference of the North American
  Chapter of the Association for Computational Linguistics: Human Language
  Technologies},  74--84.

\bibitem[\protect\citeauthoryear{Shen, Wang, and Han}{2015}]{shen2015entity}
Shen, W.; Wang, J.; and Han, J.
\newblock 2015.
\newblock Entity linking with a knowledge base: Issues, techniques, and
  solutions.
\newblock {\em IEEE Transactions on Knowledge and Data Engineering}
  27(2):443--460.

\bibitem[\protect\citeauthoryear{Srivastava \bgroup et al\mbox.\egroup
  }{2014}]{srivastava2014dropout}
Srivastava, N.; Hinton, G.; Krizhevsky, A.; Sutskever, I.; and Salakhutdinov,
  R.
\newblock 2014.
\newblock Dropout: a simple way to prevent neural networks from overfitting.
\newblock {\em The Journal of Machine Learning Research} 15(1):1929--1958.

\bibitem[\protect\citeauthoryear{Strubell \bgroup et al\mbox.\egroup
  }{2017}]{strubell2017fast}
Strubell, E.; Verga, P.; Belanger, D.; and McCallum, A.
\newblock 2017.
\newblock Fast and accurate entity recognition with iterated dilated
  convolutions.
\newblock In {\em Proceedings of the 2017 Conference on Empirical Methods in
  Natural Language Processing},  2670--2680.

\bibitem[\protect\citeauthoryear{Surdeanu \bgroup et al\mbox.\egroup
  }{2012}]{surdeanu2012multi}
Surdeanu, M.; Tibshirani, J.; Nallapati, R.; and Manning, C.~D.
\newblock 2012.
\newblock Multi-instance multi-label learning for relation extraction.
\newblock In {\em Proceedings of the 2012 joint conference on empirical methods
  in natural language processing and computational natural language learning},
  455--465.
\newblock Association for Computational Linguistics.

\bibitem[\protect\citeauthoryear{Vaswani \bgroup et al\mbox.\egroup
  }{2017}]{vaswani2017attention}
Vaswani, A.; Shazeer, N.; Parmar, N.; Uszkoreit, J.; Jones, L.; Gomez, A.~N.;
  Kaiser, {\L}.; and Polosukhin, I.
\newblock 2017.
\newblock Attention is all you need.
\newblock In {\em Advances in Neural Information Processing Systems},
  5998--6008.

\bibitem[\protect\citeauthoryear{Verga, Neelakantan, and
  McCallum}{2017}]{verga2017generalizing}
Verga, P.; Neelakantan, A.; and McCallum, A.
\newblock 2017.
\newblock Generalizing to unseen entities and entity pairs with row-less
  universal schema.
\newblock In {\em Proceedings of the 15th Conference of the European Chapter of
  the Association for Computational Linguistics}, volume~1,  613--622.

\bibitem[\protect\citeauthoryear{Verga, Strubell, and
  McCallum}{2018}]{naacl18-verga}
Verga, P.; Strubell, E.; and McCallum, A.
\newblock 2018.
\newblock {Simultaneously Self-attending to All Mentions for Full-Abstract
  Biological Relation Extraction}.
\newblock In {\em {Annual Conference of the North American Chapter of the
  Association for Computational Linguistics: Human Language Technologies (NAACL
  HLT)}}.

\bibitem[\protect\citeauthoryear{Wang \bgroup et al\mbox.\egroup
  }{2016}]{wang2016relation}
Wang, L.; Cao, Z.; de~Melo, G.; and Liu, Z.
\newblock 2016.
\newblock Relation classification via multi-level attention cnns.
\newblock In {\em Proceedings of the 54th Annual Meeting of the Association for
  Computational Linguistics}, volume~1,  1298--1307.

\bibitem[\protect\citeauthoryear{Wei \bgroup et al\mbox.\egroup
  }{2015}]{wei2015overview}
Wei, C.-H.; Peng, Y.; Leaman, R.; Davis, A.~P.; Mattingly, C.~J.; Li, J.;
  Wiegers, T.~C.; and Lu, Z.
\newblock 2015.
\newblock Overview of the biocreative v chemical disease relation (cdr) task.
\newblock In {\em Proceedings of the fifth BioCreative challenge evaluation
  workshop},  154--166.

\bibitem[\protect\citeauthoryear{Wei \bgroup et al\mbox.\egroup
  }{2016}]{wei2016assessing}
Wei, C.-H.; Peng, Y.; Leaman, R.; Davis, A.~P.; Mattingly, C.~J.; Li, J.;
  Wiegers, T.~C.; and Lu, Z.
\newblock 2016.
\newblock Assessing the state of the art in biomedical relation extraction:
  overview of the biocreative v chemical-disease relation (cdr) task.
\newblock {\em Database} 2016.

\bibitem[\protect\citeauthoryear{Wei, Kao, and Lu}{2013}]{wei2013pubtator}
Wei, C.-H.; Kao, H.-Y.; and Lu, Z.
\newblock 2013.
\newblock Pubtator: a web-based text mining tool for assisting biocuration.
\newblock {\em Nucleic acids research} 41(W1):W518--W522.

\bibitem[\protect\citeauthoryear{Wei, Kao, and Lu}{2015}]{wei2015gnormplus}
Wei, C.-H.; Kao, H.-Y.; and Lu, Z.
\newblock 2015.
\newblock Gnormplus: an integrative approach for tagging genes, gene families,
  and protein domains.
\newblock {\em BioMed research international} 2015.

\bibitem[\protect\citeauthoryear{Xie \bgroup et al\mbox.\egroup
  }{2016}]{xie2016representation}
Xie, R.; Liu, Z.; Jia, J.; Luan, H.; and Sun, M.
\newblock 2016.
\newblock Representation learning of knowledge graphs with entity descriptions.
\newblock In {\em Thirtieth AAAI Conference on Artificial Intelligence}.

\bibitem[\protect\citeauthoryear{Yang \bgroup et al\mbox.\egroup
  }{2014}]{yang2014embedding}
Yang, B.; Yih, W.-t.; He, X.; Gao, J.; and Deng, L.
\newblock 2014.
\newblock Embedding entities and relations for learning and inference in
  knowledge bases.
\newblock {\em arXiv preprint arXiv:1412.6575}.

\end{thebibliography}
